\documentclass[letterpaper, 10 pt, conference]{ieeeconf}  

\IEEEoverridecommandlockouts                              

\usepackage{subfigure}
\usepackage[OT1]{fontenc} 
\usepackage{graphicx} 
\usepackage{multirow}
\usepackage{algorithm}
\usepackage{algpseudocode}
\usepackage{amsmath}
\usepackage{stfloats}
\usepackage[colorlinks=true,linkcolor=blue]{hyperref}

\title{\LARGE \bf
Off-Road Drivable Area Extraction Using 3D LiDAR Data
}

\author{\authorblockN
	{Biao Gao\authorrefmark{1},
		Anran Xu\authorrefmark{1}, 
		Yancheng Pan\authorrefmark{1},
		Xijun Zhao\authorrefmark{2},
		Wen Yao\authorrefmark{2},
		Huijing Zhao\authorrefmark{1}}
	\authorblockA{\authorrefmark{1}Peking University, Beijing, China}
	\authorblockA{\authorrefmark{2}China North Vehicle Research Institute, Beijing, China}
	\thanks{This work is partially supported by the NSFC Grants 61573027. B. Gao, A. Xu, Y. Pan and H. Zhao are with the Peking University, with the Key Laboratory of Machine Perception (MOE), and also with the School of Electronics Engineering and Computer Science, Beijing 100871, China. X. Zhao and W. Yao are with China North Vehicle Research Institute, Beijing, China. Correspondence: H. Zhao, zhaohj@cis.pku.edu.cn.}}

\begin{document}

\bibliographystyle{unsrt}
\maketitle

\begin{abstract}
	
We propose a method for off-road drivable area extraction using 3D LiDAR data with the goal of autonomous driving application. A specific deep learning framework is designed to deal with the ambiguous area, which is one of the main challenges in the off-road environment. To reduce the considerable demand for human-annotated data for network training, we utilize the information from vast quantities of vehicle paths and auto-generated obstacle labels. Using these auto-generated annotations, the proposed network can be trained using weakly supervised or semi-supervised methods, which can achieve better performance with fewer human annotations. The experiments on our dataset illustrate the reasonability of our framework and the validity of our weakly  and semi-supervised methods.

\end{abstract}

\section{INTRODUCTION}

Recent advances in self-driving vehicles have been very impressive. Drivable area extraction is a key technology in this domain and a prerequisite for safe and reliable autonomous driving\cite{BarHillel2014}.
Currently, the more mature techniques have mainly been designed for urban structured road environments\cite{He2004}\cite{Alvarez2011}
, but few studies have focused on off-road environments. In off-road environments, there are no structured features such as traffic lanes, paved road surfaces or guardrails. The off-road drivable area usually has ambiguous margins, various textures and complex features, which creates considerable challenges extracting the drivable area.
As a result, the algorithms designed for the urban environment are difficult to apply directly to the off-road environment. 

Cameras and LiDAR are two main sensors that provide input data for drivable area extraction tasks. There are many camera-based methods that have already been applied in off-road environments\cite{Mei2018}.
However, the color or texture features they used are not robust enough in diverse illumination and weather conditions. The lack of 3D information limits the performance and adaptability of these methods in different scenes as well\cite{Xiao2015}.
LiDAR has been widely used in self-driving systems because of its advantages in collecting 3D point cloud data directly. There are some LiDAR-based methods that depend on data segmentation and rule/threshold-based methods to extract the drivable area\cite{Zhang2010}.
However, these methods rely heavily on human-designed features and presupposed thresholds, and they usually have poor scene adaptability. 

\begin{figure}[t]
	\centering
	\includegraphics[scale=0.17]{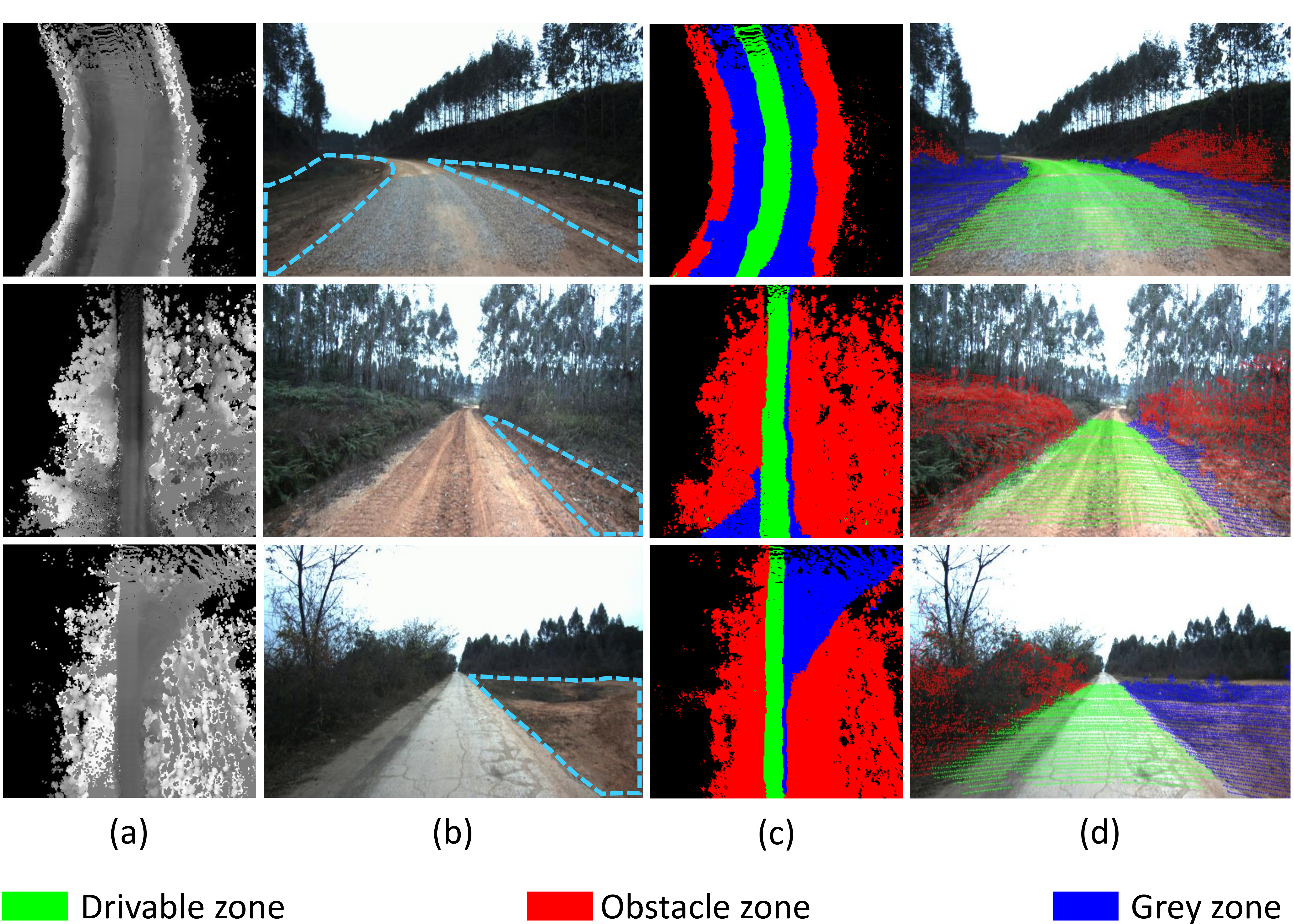}
	\caption{The ambiguities in off-road drivable area extraction. (a) Input LiDAR data in bird's-eye view (the ego-car is in the center of images with an upward heading). (b) Image reference of input data. (c) Human-annotated labels. (d) Human-annotated labels' projection on image}
	\label{fig:example}
	\vspace{-3mm}
\end{figure}

In this work, we focus on off-road drivable area extraction using 3D LiDAR data. To illustrate the main challenges of the off-road scene, we use light blue polygons to include some typical ambiguous areas in column (b) of Figure.\ref{fig:example}. A human driver would not enjoy driving in these areas because of their higher traversability costs, but they are technically drivable to some degree. It is unreasonable to simply label these ambiguous areas as either drivable zones or the obstacle zones, so they are called grey zones in this paper.

We propose a deep learning method for drivable area extraction using 3D LiDAR data specific to the off-road environment. Compared with traditional human-designed features, the proposed method can autonomously learn features of the drivable zone from the labelled data. Additionally, it is suitable for weakly supervised and semi-supervised learning. By combining the features from the vehicle paths and auto-generated vertical obstacles, our method can significantly decrease the demand for human annotation in the neural network training. The experimental results prove the validity of our proposed method.

This paper is organized as follows. First, the related works are briefly introduced in Section.\ref{sec:relatedworks}. Section.\ref{sec:method} introduces the methodology in detail. Section.\ref{sec:implementationdetails} and Section.\ref{sec:experiment} show some implementation details and the experimental results. Finally, we draw conclusions in Section.\ref{sec:conlusion}.

\section{RELATED WORKS}	\label{sec:relatedworks}
Cameras are one of the most important sensors in the road/drivable area extraction tasks for autonomous vehicles. Some camera-based methods depend on the assumption of global road priors such as road boundaries\cite{Yuan2015},
traffic lanes\cite{Aly2014}\cite{ZuWhanKim2008} or
vanish points\cite{Alvarez2014}\cite{Audibert2010}
. Some other studies do not rely on these assumptions but view the drivable area extraction as a segmentation of road and non-road regions\cite{Mei2018}\cite{Zhou2010}. Furthermore, some stereo camera based approaches\cite{rankin2009stereo}\cite{broggi2013terrain} make use of depth information to help off-road drivable area extraction.
Despite achieving good performance, camera-based methods are easily affected by changing illumination. 
The LiDAR-based methods can address this weakness, and the higher precision 3D information can be conveniently used to extract the road boundaries\cite{Zhang2010}\cite{Wijesoma2004}\cite{Zhang2018} or fit the road plane\cite{Asvadi2016}\cite{Hu2014}. For the different characteristics of the two sensors, LiDAR-camera fusion becomes a natural solution. For example, Dahlkamp\cite{Stanford2006} identified a nearby drivable area by LiDAR and used it to train an image-based classifier for far-range drivable area detection.

The existing drivable area extraction methods are mostly designed for urban environment, but the problem in off-road environments is quite different. One fundamental problem is the ambiguous definition of the drivable area, as shown in Figure.\ref{fig:example}. Many studies have proposed similar concepts for off-road scenes from different perspectives, such as traversability analysis\cite{Suger2015} and drivable corridors\cite{Nefian2006}.
Despite some methods having already been implemented in off-road environments\cite{Audibert2010}\cite{Stanford2006}\cite{Nefian2006}, they still have some limits. These methods usually focus on the mechanically drivable area but seldom distinguish whether these areas are likely to be chosen by a human driver, which makes great sense for autonomous vehicles.

Recently, many deep learning methods have achieved impressive results on related tasks\cite{barnes2017find}\cite{Han2018}\cite{Oliveira2016}\cite{chen2015deepdriving}\cite{Bellone2018}. Compared to the traditional methods, deep networks can learn high-level semantic features directly from the data, which usually performs better than human-designed features.
However, the deep learning methods usually rely on large human-annotated datasets such as KITTI\cite{geiger2013vision} and Cityscapes\cite{cordts2016cityscapes}. For off-road environments, there are few widely used datasets for the drivable area extraction due to the ambiguous problem definition. 
To reduce the demand for human annotation, some studies have used a simulator to access endless data for training\cite{chen2015deepdriving}. In addition, other studies have focused on weakly supervised\cite{barnes2017find} or semi-supervised methods\cite{Suger2015}, attempting to use auto-generated weak labels as substitution, which can be easily accessed.

This work focuses on drivable area extraction in off-road environments. We propose a LiDAR-based deep learning framework specific to the ambiguities in this task. To reduce the demand for human-annotated datasets, we also propose weakly supervised and semi-supervised methods to learn features from auto-generated labels.

\section{METHODOLOGY}	\label{sec:method}

\subsection{Problem Definition}

We aggregate a few frames' point clouds to obtain a dense bird's-eye view height map $X=\{x_{j,k}\}_{0\le j<H,0\le k<W}$, which is used as our input data format. The input height map is the size of $H\times W$ and each pixel $x_{j,k}$ represents the physical height of pixel $(j,k)$. The car is in the center of $X$ with an upward heading. The input examples can be seen in Figure.\ref{fig:example}(a).

Different from the well-defined road borders in structured urban environments, the main peculiarity in off-road environments is the ambiguous area beside the road margin, which is called the grey zone. To distinguish it with others, we let $Label Set=\{unknown,drivable\ zone,obstacle\ zone,grey\ zone\}$, and we use $G=\{g_{j,k}\}_{0\le j<H,0\le k<W}$ to denote the human annotated ground truth, where $g_{j,k}\in LabelSet$.

The original output of our proposed framework is a cost map $C=\{c_{j,k}\}_{0\le j<H,0\le k<W}$, where each $c_{j,k}\in [0,1]$ evaluates the traversability cost of pixel $(j,k)$. Our proposed framework learns a mapping from input $X$ to cost map $C$.

\vspace{-2mm}
\begin{equation}
f_\theta^*:x_{j,k}\rightarrow c_{j,k} \in [0,1]
\end{equation}

For conveniently comparing with the human-annotated ground truth $G$ and other baseline methods, we use Equation (\ref{equ:Cost}) to get label $Y=y_{j,k}\in LabelSet$.


Therefore, the problem in this work can be formulated as learning a multi-class classifier $f_\theta$ that maps input $x_{j,k}$ to a label $y_{j,k}\in LabelSet$.

\subsection{Network Architecture}
Due to the ambiguity of the grey zone, we hold the view that classifying it as an independent label from the others is not reasonable enough. In some cases, the grey zone is technically drivable but not human-desired, which is very close to or even overlaps with the drivable zone in the feature space. In other cases, the grey zone may have a higher traversability cost than the common drivable zone, which is closer to or even overlaps the obstacle zone in feature space. As a result, viewing the ambiguous grey zone as an independent label in the training process will cause confusion for the deep learning model, and the experimental results in Section.\ref{sec:experiment} give evidence of this viewpoint.

\begin{figure}[b]
	\vspace{-2mm}
	\centering
	\includegraphics[scale=0.22]{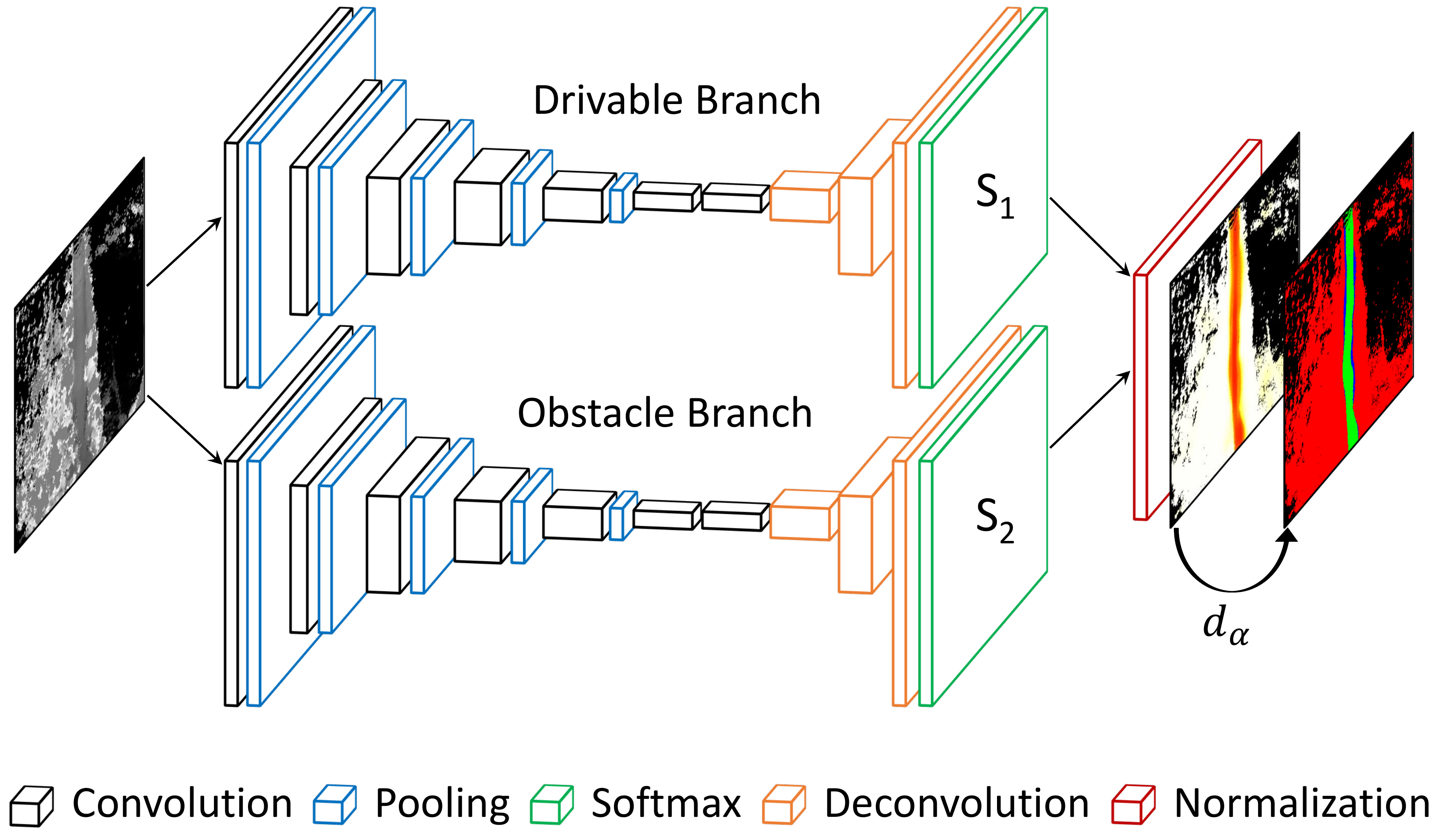}
	\caption{The proposed network architecture}
	\label{fig:network}
\end{figure}

\begin{figure*}[ht]
	\centering
	\includegraphics[scale=0.37]{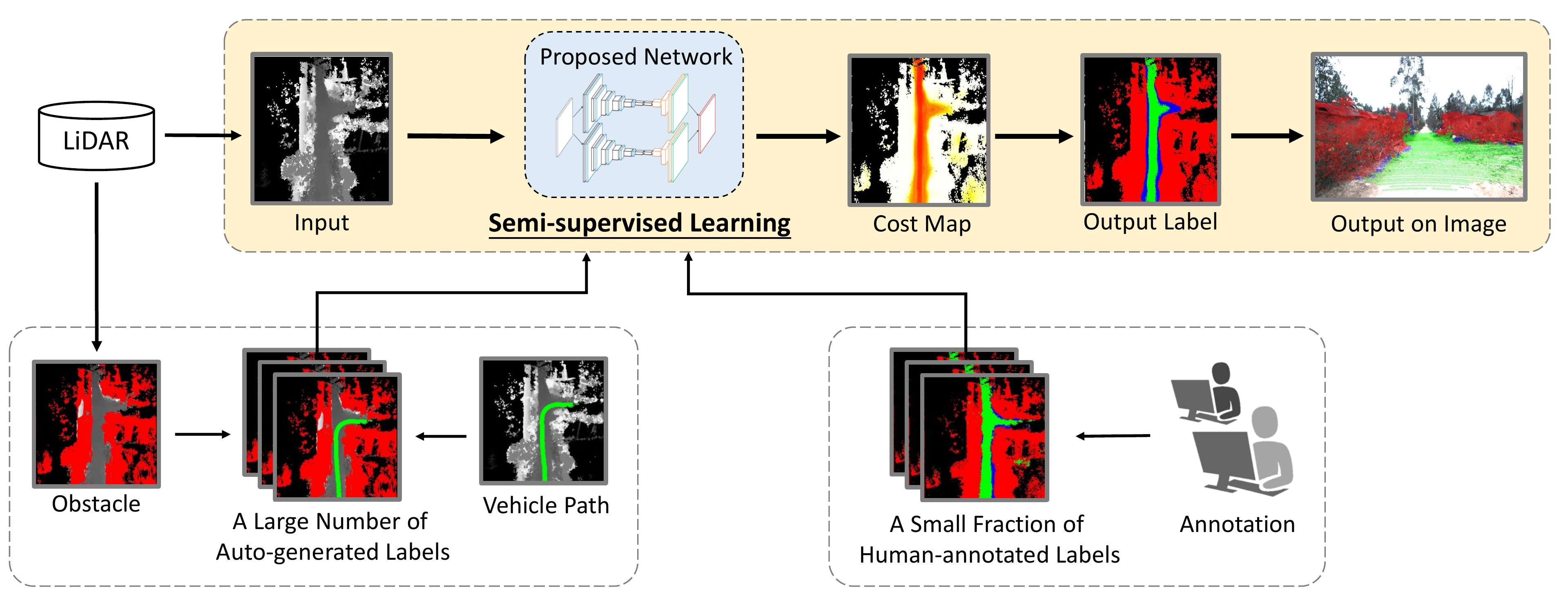}
	\caption{Overview of the proposed off-road drivable area extraction framework}
	\label{fig:framework}
	\vspace{-4mm}
\end{figure*}

Therefore, we can assume that the grey zone samples are distributed between the drivable samples and the obstacle samples in the feature space. 
The key idea of our proposed method is learning two classification surfaces to separate the grey zone samples. One classification surface is used to separate the drivable zone samples from the other samples. The other classification surface is used to separate the obstacle zone samples. As a result, we can evaluate the traversability cost of a sample by its feature distance to the two surfaces.                                                                                                                                                                                                                                 

As shown in Figure.\ref{fig:network}, the proposed network has two branches for learning the two classification surfaces mentioned above. Both of these branches are designed according to the common VGG-based fully convolutional network. The difference is that the last layer does not output the discrete labels but the probabilistic predictions from the last softmax layer. We denote them as $S_1$ and $S_2$, which are in the range of $[0,1]$.

Each branch is trained end-to-end guided by the following cross-entropy loss function:

\vspace{-4mm}
\begin{equation}
\label{equ:loss0}
\resizebox{.91\hsize}{!}{
	$L^{br}(X;\Theta^{br})=-\sum_{i}{y_i^{br} \log{P(y_i^{br}|\Theta^{br})}}, br \in \{dri, obs\}$
}
\end{equation}

where $br\in\{dri, obs\}$ is the name of the network's branch. $P(y_i^{br}|\Theta^{br})$ is the probability that  pixel $i$ is predicted as label $y_i^{br}$ with the network parameters $\Theta^{br}$. We use $dri,obs$ and $gre$ to represent the $drivable, obstacle$ and $grey\ zone$.

When training the network, a different label, $y_i^{br}$, is used in the two branches. 

\vspace{-2mm}
\begin{equation}
y_i^{br}= 
\left\{
\begin{array}{ll}
\Psi(\vec{br}), &\quad if \ g_i = gre \\ 
\vec{g_i}, &\quad if \ g_i \in \{dri,obs \} \\
\end{array}
\right.
\end{equation}
\vspace{-3mm}

where $\vec{br}$ is the one-hot vector of label $br\in\{dri, obs\}$. $\Psi(\vec{dri})=\vec{obs}$ and $\Psi(\vec{obs})=\vec{dri}$. Concretely, when training the drivable branch, we replace all the pixels satisfied $g_i=gre$ with label $obs$ to obtain $y_i^{dri}$. When training the obstacle branch, we replace all the pixels satisfied $g_i=gre$ with label $dir$ to obtain $y_i^{obs}$.

We use the following regulation to calculate the traversability cost map $C$ and the discrete label:

\vspace{-3mm}
\begin{equation}
\label{equ:Cost}
C= 
\left\{
\begin{array}{ccc}
S_1, &\  if \ S_1>\alpha_1\ and\ S_2<\alpha_2\\ 
1-S_2, &\ if \ S_2>\alpha_2\ and\ S_1<\alpha_1\\
\frac{1-S_2}{1-S_1 + 1-S_2}, &\ otherwise
\end{array}
\right.
\end{equation}

where $\alpha_1$ and $\alpha_2$ are hyper-parameters, the pixels satisfying the first condition are labelled as $y_{j,k}=dri$, the second is $y_{j,k}=obs$ and the last is $y_{j,k}=gre$.

\subsection{Weakly and Semi-supervised Learning}

To reduce the demand for the high-cost human-annotated data, we propose a semi-supervised learning method shown in Figure.\ref{fig:framework}. For our weakly supervised method, no human-annotated labels are used for training. In addition, for our semi-supervised method, this framework can combine a large number of auto-generated labels (see Section.\ref{sec:autolabel} for more details) and only a small fraction of human-annotated labels to train the network. Except for the numerous auto-generated labels, this framework is almost the same as the fully supervised framework. Our proposed network receives the LiDAR-based height maps as the inputs and outputs the traversability cost map for each pixel. For the convenience of evaluation, the cost map is discretized to the 3-class result and projected on the image for visualization.

For human-annotated data $X_h$, we use the loss function Equation (\ref{equ:loss0}) for training. For data $X_w$ with only auto-generated weak labels, we define the loss  $L_{semi}^{br}$ in branch $br$ as below:

\vspace{-4mm}
\begin{equation}
\label{equ:loss1}
L^{br}_{semi}(X_w;\Theta^{br})=-\lambda \sum_{i}{\widetilde{y_i^{br}} \log{P(\widetilde{y_i^{br}}|\Theta^{br})}}
\end{equation}

\vspace{-2mm}
where $\lambda$ is a regularization weight. $\widetilde{y_i^{br}}$ is used for weakly and semi-supervised training, which has a similar definition as $y_i^{br}$ except for the pixels without auto-generated labels $\widetilde{g_i}$. $unk$ means the unknown zone. $g_i\neq unk$ represents the pixels without LiDAR observation, so they are labeled as unknown.

\begin{table*}
	\vspace{2mm}
	\caption{Evaluation Measures}
	\label{tab:evaluation}
	\centering
	\renewcommand{\arraystretch}{1.5}
	\begin{tabular}{cclccl}
		\hline
		\multicolumn{3}{c|}{Drivable Zone}                                                                                                    & \multicolumn{3}{c}{Obstacle Zone}                                                                                                  \\ \hline
		Definition                                            & \multicolumn{2}{c|}{Explanation}                                               & Definition                                          & \multicolumn{2}{c}{Explanation}                                               \\ \hline
		$Q_1={TP(G_{dri})}/{\Arrowvert Y_{dri} \Arrowvert}$   & \multicolumn{2}{c|}{$TP(G_{dri})=\Arrowvert G_{dri}\cap Y_{dri} \Arrowvert$}   & $Q_1={TP(G_{obs})}/{\Arrowvert Y_{obs} \Arrowvert}$ & \multicolumn{2}{c}{$TP(G_{obs})=\Arrowvert G_{obs}\cap Y_{obs} \Arrowvert$}   \\
		$Q_2={TP(G_{dri})}/{\Arrowvert G_{dri} \Arrowvert}$   & \multicolumn{2}{c|}{$TP(G_{dri})=\Arrowvert G_{dri}\cap Y_{dri} \Arrowvert$}   & $Q_2={TP(G_{obs})}/{\Arrowvert G_{obs} \Arrowvert}$ & \multicolumn{2}{c}{$TP(G_{obs})=\Arrowvert G_{obs}\cap Y_{obs} \Arrowvert$}   \\
		$Q_3={TP(VP_{dri})}/{\Arrowvert VP_{dri} \Arrowvert}$ & \multicolumn{2}{c|}{$TP(VP_{dri})=\Arrowvert VP_{dri}\cap Y_{dri} \Arrowvert$} & /                                                   & \multicolumn{2}{c}{/}                                                         \\
		$F_1={2Q_1Q_2}/{(Q_1+Q_2)}$                           & \multicolumn{2}{c|}{$F_1$ measure}                                            & $F_1={2Q_1Q_2}/{(Q_1+Q_2)}$                         & \multicolumn{2}{c}{$F_1$ measure}                                             \\ 
		\hline
	\end{tabular}
	\renewcommand{\arraystretch}{2.3}
	\begin{tabular}{llllll}
		\textbf{dri}: Drivable zone                 & \textbf{obs}: Obstacle zone              & \textbf{G}: Ground truth                            & \textbf{Y}: Predicted label & \textbf{VP}: Vehicle path & $\Arrowvert \textbf{X}\Arrowvert$: Pixel number in X \\
	\end{tabular}
	\vspace{-4mm}
\end{table*}

\vspace{-5mm}
\begin{equation}
\widetilde{y_i^{br}}= 
\left\{
\begin{array}{ll}
\Psi(\vec{br}), &\quad if \ \widetilde{g_i} = unk\ \&\ g_i \neq unk \\ 
y_i^{br}, &\quad otherwise \\
\end{array}
\right.
\end{equation}

When training the network with human-annotated labels and auto-generated labels simultaneously, the Equation (\ref{equ:loss0}) and Equation (\ref{equ:loss1}) are combined.

\vspace{-4mm}
\begin{equation}
\label{equ:loss2}
L^{br}(X_h,X_w;\Theta^{br})=L^{br}(X_h;\Theta^{br})+L^{br}_{semi}(X_w;\Theta^{br})
\end{equation}

In each training batch, the human-annotated and auto-generated data are randomly fed to the model, and their mean loss is calculated for backpropagation. If only the auto-generated labels are used for training (weakly supervised learning), Equation (\ref{equ:loss1}) is the loss function.

\section{IMPLEMENTATION DETAILS} \label{sec:implementationdetails}

\subsection{Automatic Labelling}	\label{sec:autolabel}

As illustrated in Figure.\ref{fig:framework}, we use the recorded data from the data collection car to achieve the auto-generated labels.
We follow the rules-based region growing method described in Algorithm\ref{alg:rg} to generate the vertical obstacles from the LiDAR data as the weak obstacle zone labels. We only need to set a loose threshold for region growing, and we can obtain a relatively strict vertical obstacle area.

In addition, we assume that the vehicle path chosen by the human driver must belong to the drivable zone. Therefore, we project the data collection car's GPS trajectories with the same width as the car to the input height map, and they are labelled as the weakly drivable zone.

\begin{algorithm}	
	\caption{Region Grow}
	\label{alg:rg}
	\begin{algorithmic}[1]
		\Require Height map $X$, Height threshold $T_h$, Angle threshold $T_a$, Initial road height interval $[T_r, T_r']$
		\Ensure Drivable region set $S_d$, Obstacle region set $S_o$
		\State Initialize: waiting list $Q=\emptyset,S_d=\emptyset,S_o=\emptyset$
		\State Height difference: $\Delta H$
		\State Angle difference: $\Delta A$
		\ForAll {$x_i$ \textbf{in} $X$}
		\If {$x_i \in [T_r, T_r']$}
		\State $Q \gets Q \cup x_i$, $S_d \gets S_d \cup x_i$
		\EndIf
		\EndFor
		\While {$Q \neq \emptyset$}
		\ForAll {$x_j \in NeighbourSet(x_i)$}
		\If {$\Delta H(x_i,x_j) < T_h$ \textbf{and} $\Delta A(x_i,x_j) < T_a$}
		\State $Q \gets Q \cup x_j$, $S_d \gets S_d \cup x_j$
		\Else
		\State $S_o \gets S_o \cup x_j$
		\EndIf
		\EndFor
		\State $Q \gets Q-x_i$
		\EndWhile
	\end{algorithmic}
\end{algorithm}

\subsection{Training Setup}
The training process and experiments are conducted on a NVIDIA TitanX GPU. The network is trained with the Adam optimizer with the learning rate 1e-4 and the batch size of 16. Data augmentation is applied mainly for image rotation because the vehicle seldom makes a turn. In semi-supervised learning loss, the regularization weight $\lambda$ is usually less than 1, such as $0.4$. When converting the cost map into discrete labels, we set the thresholds $\alpha_1$ and $\alpha_2$ to 0.5 for an SUV. Actually, they can be changed based on the through capacity of a vehicle. 

\subsection{Evaluation Measures}	\label{sec:eval}

To evaluate the quantitative performance of different algorithms in the off-road environment, we design some evaluation measures and present them in Table.\ref{tab:evaluation}. 

Due to the ambiguous definition of the grey zone, we do not evaluate the performance on the grey zone samples directly, but only the drivable zone and obstacle zone samples.

\subsubsection{Precision}
We define $Q_1$ to evaluate the precision performance. For the drivable zone, $Q_1={TP(G_{dri})}/{\Arrowvert Y_{dri} \Arrowvert}$. ${\Arrowvert Y_{dri} \Arrowvert}$ represents the number of pixels predicted as the drivable zone. $Q_1$ measures the percentage of extracted drivable pixels that are the actual drivable zone in the ground truth. For the obstacle zone, $Q_1$ measures the percentage of extracted obstacle pixels that belong to the obstacle zone in human annotations.
\subsubsection{Recall}
We define $Q_2$ to evaluate the recall. For the drivable zone, $Q_2={TP(G_{dri})}/{\Arrowvert G_{dri} \Arrowvert}$, where ${\Arrowvert G_{dri} \Arrowvert}$ represents the number of pixels in the drivable zone in the ground truth. $Q_2$ measures the percentage of the ground truth drivable zone extracted by the proposed method. For the obstacle zone, $Q_2$ is defined in a similar fashion.
\subsubsection{Accuracy}
$Q_3$ is defined only for the drivable zone, which measures the percentage of the vehicle path extracted as the drivable zone. We believe that the vehicle paths chosen by the human driver must be the area with a relatively low traversability cost. Therefore, we design $Q_3$ to encourage extracting the vehicle path pixels as the drivable zone.
\subsubsection{$F_1$ Measure}
Finally, the $F_1$ measure is a widely used indicator that considers both the precision and the recall. The $F_1$ measure is the harmonic average of the precision ($Q_1$) and recall ($Q_2$). 

In this work, there are some methods that tend to extract the drivable zone with a narrow-width, which is similar to a vehicle path. This leads to very high performance for precision ($Q_1$) but lower recall ($Q_2$). In addition, there are some other methods that tend to extract a wider drivable zone, which leads to higher recall ($Q_2$) but lower precision ($Q_1$). In these cases, the $F_1$ measure is considered the most important indicator for evaluating the method’s performance.

\begin{table*}[b]
	\vspace{-2mm}
	\caption{quantitative evaluation of different methods}
	\label{tab:all_result}
	\centering
	\renewcommand{\arraystretch}{1.5}
	\begin{tabular}{c|cccc|ccc}
		\hline
		& \multicolumn{4}{c|}{Drivable zone}                                & \multicolumn{3}{c}{Obstacle zone}               \\ \cline{2-8} 
		& $Q_1$ (PRE)    & $Q_2$ (REC)    & $Q_3$ (ACC)    & $F_1$          & $Q_1$ (PRE)    & $Q_2$ (REC)    & $F_1$          \\ \hline
		3-class FCN (fully sup.) & 74.93          & 82.99          & \textbf{98.92} & 78.75          & 94.36          & \textbf{98.44} & 96.36          \\
		Ours (fully sup.)        & \textbf{76.01} & \textbf{86.72} & 98.09          & \textbf{81.01} & \textbf{96.20} & 96.75          & \textbf{96.47} \\ \hline
		RG-FCN (weakly sup.)     & 59.78          & {79.15}        & 93.16          & 68.11          & 94.46          & {95.38}        & 94.92          \\
		Oxford PP (weakly sup.)  & \textbf{97.00} & 47.38          & 83.71          & 63.66          & \textbf{98.40} & 89.84          & 93.93          \\
		Ours (weakly sup.)       & 72.38          & 78.83          & {95.21}        & {75.47}        & {96.31}        & 94.84          & {95.57}        \\
		Ours (semi-sup.)         & {81.73}        & \textbf{81.73} & \textbf{96.24} & \textbf{81.73} & 95.60          & \textbf{97.38} & \textbf{96.49} \\ \hline
	\end{tabular}
\end{table*}

\section{EXPERIMENTAL RESULTS}	\label{sec:experiment}

\subsection{Dataset}
We build a typical off-road dataset using our data collection vehicle, which is equipped with a Velodyne HDL-64 LiDAR, a front-view monocular camera and a GPS/IMU system. To collect the input data $X$, We project point clouds captured by the LiDAR into a birds-eye view. The position and posture information captured by the GPS/IMU system is required during the projection process. In addition, we use the GPS/IMU system to record the vehicle trajectory for automatically labelling the vehicle path. We emphasize that the camera data are only used for visualization.

During the data collection processing, the vehicle is driven by a human driver and all kinds of data are time-synchronized. The input height map $X$ is in the size of $300\times 300$ with 0.2 metres pixel size. The height value of each pixel is linearly projected to an integer in $[0,255]$.

The whole dataset contains 1961 frames of data. The driving distance is approximately 785 meters. We choose $60\%$ frames for model training, $15\%$ frames for validation and $25\%$ frames for testing.

\subsection{Proposed Method Results}
To evaluate our proposed method's performance, we first compare it with a fully supervised method to prove the reasonability of our model design for the grey zone. Another advantage of our model is that it is also suitable for weakly and semi-supervised learning. Therefore, we compare our weakly and semi-supervised results with other baselines.

\begin{figure*}[ht]
	\centering
	\includegraphics[scale=0.15]{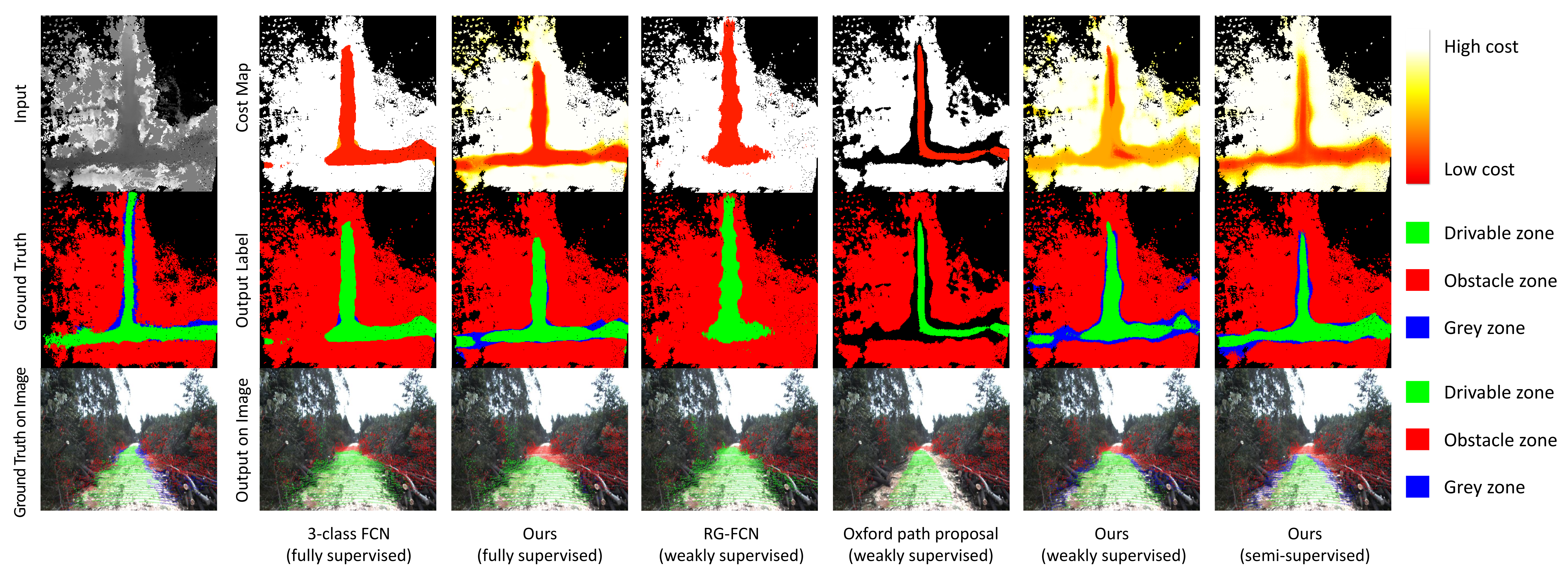}
	\caption{Qualitative results at crossroad scene.}
	\label{fig:cross_road}
	\vspace{-3mm}
\end{figure*}

\begin{figure*}[ht]
	\centering
	\includegraphics[scale=0.15]{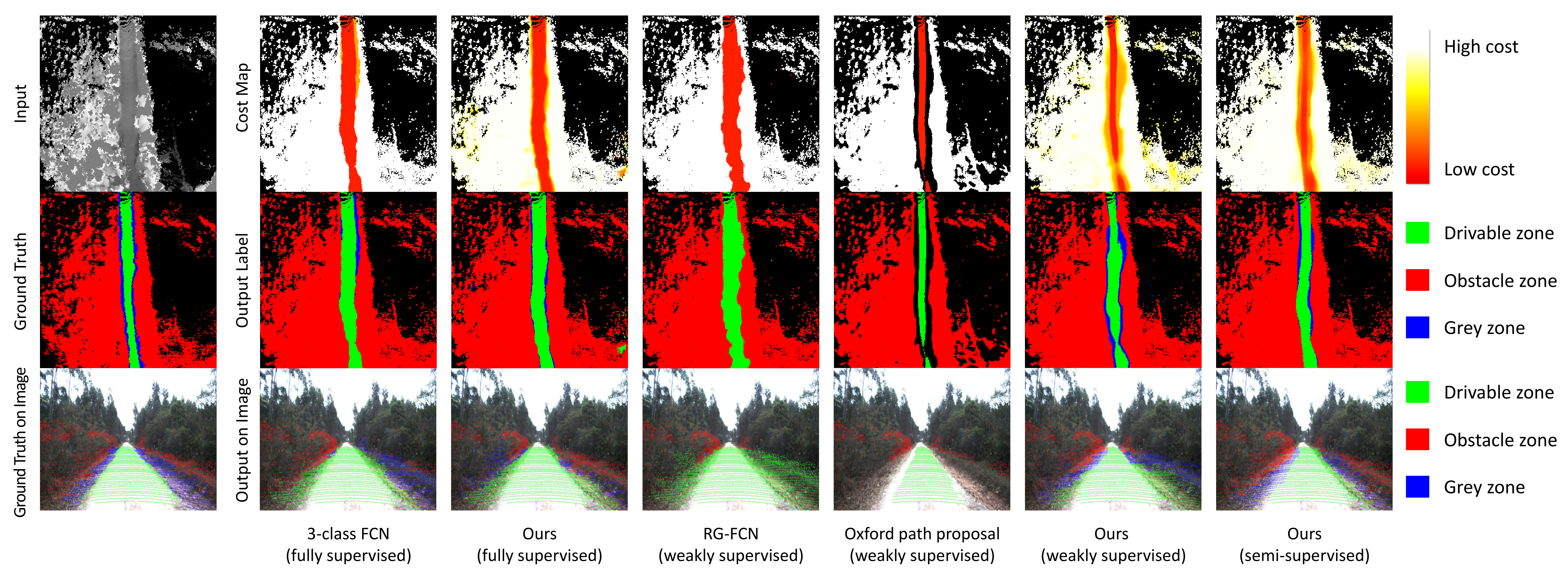}
	\caption{Qualitative results at straight road scene.}
	\label{fig:straight_road}
	\vspace{-3mm}
\end{figure*}

Figure.\ref{fig:cross_road} and Figure.\ref{fig:straight_road} show the qualitative test results of different methods in two typical scenes: a crossroad scene and a straight road scene. It is necessary to mention that the cost maps of other baseline methods are directly remapped from their output labels, which only have 3 discrete values.

We use the evaluation measures described in Section.\ref{sec:eval} to compare the quantitative performance of different methods, which are shown in Table.\ref{tab:all_result}. 


\subsubsection{Fully Supervised Results}
Fully supervised results use the human-annotated ground truth for model training. The baseline method '3-class FCN' is based on a fully convolutional network\cite{long2015fully} with the same depth as branches in our proposed network. It treats the problem as a common 3 class classification. Our proposed fully supervised method achieves better performance than '3-class FCN' in most evaluation measures.

We can use the first three columns in Figure.\ref{fig:cross_road} for a more specific example. Our fully supervised method is more robust than the common 3-class classification method in the complex crossroad scene. The baseline '3-class FCN' misclassifies the left side crossroad as an obstacle zone and our method successfully extracts the whole structure of this crossroad.

\subsubsection{Weakly Supervised Results}
The weakly supervised method means that all data for model training are auto-generated weak labels.

We introduce two baseline methods for comparison. The first one is denoted as 'RG-FCN', which uses the traditional rule-based region growing method described in Algorithm\ref{alg:rg} to generate weak labels. These weak labels are used to train a FCN\cite{long2015fully} with cross-entropy loss function. The second baseline method 'Oxford path proposal'\cite{barnes2017find} was original designed for the task of path proposal based on image data. It achieved great performance on KITTI dataset\cite{geiger2013vision}. Due to the lack of LiDAR-based weakly supervised methods, we re-implement this method in our LiDAR-based framework as another baseline.

The qualitative visualization results are shown in the last four columns of Figure.\ref{fig:cross_road} and Figure.\ref{fig:straight_road}. From the visualization results, it is easy to find that the 'RG-FCN' method tends to extract a wider drivable zone than the ground truth. The rule-based method cannot distinguish the drivable zone and the grey zone with a few of the thresholds. The 'Oxford path proposal' results are opposite in that they have a very narrow drivable zone similar to the vehicle path. Its fundamental defect is that too many pixels between this narrow drivable zone and the obstacle zone are labelled as unknown. A large percentage of these pixels are actually drivable, and their accurate prediction makes great sense for autonomous vehicles. Compared with these two baseline methods, our weakly supervised method has obviously better performances in extracting the drivable area. In addition, our method is also as robust as the fully supervised method in the crossroad scene.

The quantitative analysis shows that our weakly supervised method can extract the drivable area more accurately than others. Despite one baseline method that has a higher precision $Q_1$, and the other method that has slightly higher recall $Q_2$, they both obtained very poor performance of the other evaluation measure. In other words, our method is the most balanced method, which obtained an $F_1$ measure that was 7.4\% higher than the 'RG-FCN' and 11.8\% higher than the 'Oxford path proposal' method.

\subsubsection{Semi-supervised Results}
The semi-supervised methods mean a proportion of human-annotated and auto-generated data are used for training at the same time. Regardless of the number of human-annotated labels, we use all weak labels for training, for their quite low generating cost.
We list the semi-supervised result using half human-annotated labels for training as representative in Table.\ref{tab:all_result}. The $F_1$ measure of our semi-supervised method (81.73\%) achieves an impressive improvement compared to other baseline methods, and it was even higher than the fully supervised result.
Our method achieves better evaluations than other methods in all measures except precision $Q_1$; the explanation of this is similar to that of the weakly supervised models. 
\begin{table*}
	\vspace{2mm}
	\caption{$F_1$ measure of different methods}
	\label{tab:semic}
	\centering
	\renewcommand{\arraystretch}{1.5}
	\begin{tabular}{c|ccccc}
		\hline
		- & 6.25\% semi-sup. & 12.5\% semi-sup. & 25\% semi-sup. & 50\% semi-sup. & fully sup.	\\
		\hline
		$F_1$ measure & 73.86 & 75.04 & 78.96 & \textbf{81.73} & 81.01 	\\
		\hline
	\end{tabular}
	\vspace{-3mm}
\end{table*}

\begin{figure}[h]
	\centering
	\includegraphics[scale=0.5]{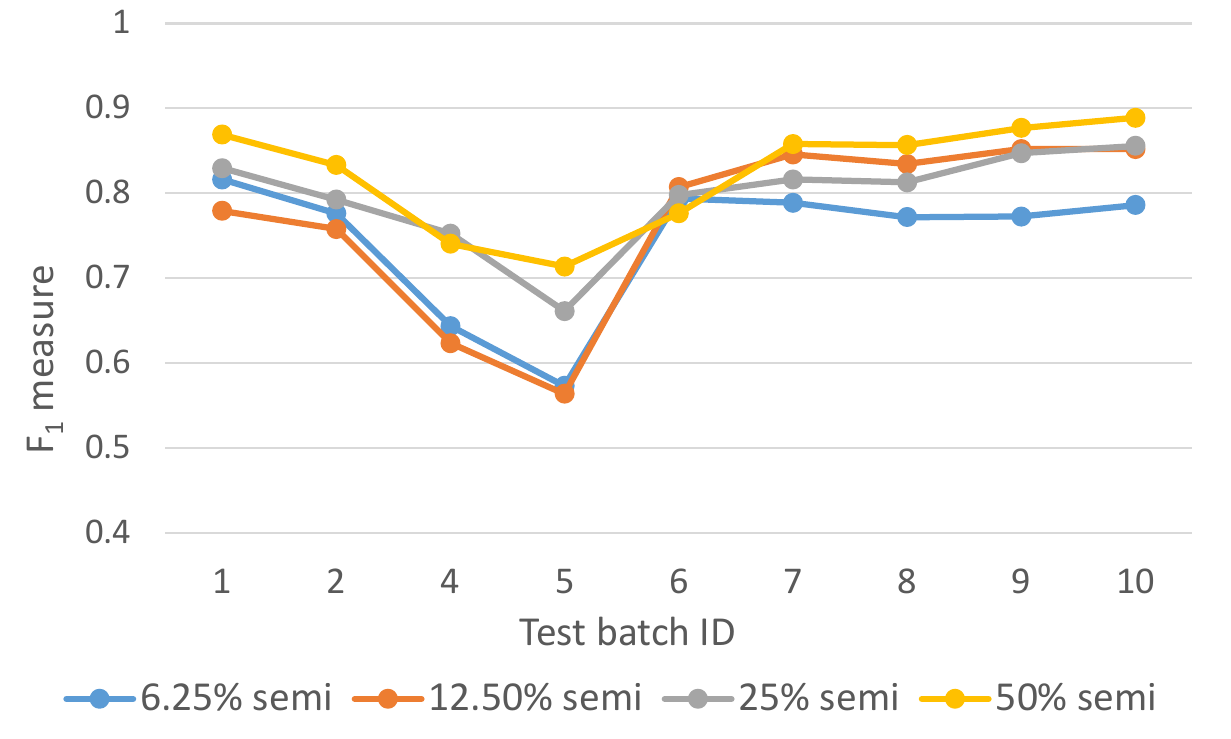}
	\caption{Detailed comparison of semi-supervised methods on the test set.}
	\label{fig:semi_curve}
	\vspace{-4mm}
\end{figure}

To explore how the ratio of human-annotated labels influences our model performance, we compare the semi-supervised methods using different ratios of human-annotated labels based on the key indicator $F_1$ measure. The detailed performance on the test set can be seen in Figure.\ref{fig:semi_curve}. We split the test set into 10 batches and evaluated on them separately. The quantitative results are shown in Table.\ref{tab:semic}. The percentage in the front of 'semi-sup' represents the ratio of human-annotated labels used for training. The $F_1$ measure of the 50\% semi-supervised version is higher than the fully supervised method, and the 25\% semi-supervised version achieves higher performance than the '3-class FCN' baseline with only a quarter of human-annotated labels. It shows that our proposed semi-supervised method can significantly reduce the demand for high-cost human annotations.

\vspace{-4mm}
\section{CONCLUSION}   \label{sec:conlusion}
In this paper, we propose a deep learning framework for off-road drivable area extraction. The proposed network structure is specifically designed for the ambiguous grey zone in the off-road environment. We also propose an automatic labelling method that generates quantities of weak labels from the vehicle's driving data collection. Our method can significantly reduce the demand for human-annotated data for the weakly and semi-supervised network training.
Importantly, it is demonstrated that the proposed semi-supervised method can achieve better performance than the fully supervised method with even fewer human-annotated labels. In this work, the camera images are only used for visualization, but they actually include many useful features for the drivable area extraction, such as the colors and textures. We plan to fuse the camera information into our framework and enhance the robustness of far-field drivable area extraction in the off-road environment.



\end{document}